\documentclass{article}

\PassOptionsToPackage{numbers, compress}{natbib}
\usepackage[preprint]{neurips_2026}

\usepackage[utf8]{inputenc}
\usepackage[T1]{fontenc}
\usepackage{hyperref}
\usepackage{url}
\usepackage{booktabs}
\usepackage{amsfonts}
\usepackage{amsmath}
\usepackage{amssymb}
\usepackage{nicefrac}
\usepackage{microtype}
\usepackage{xcolor}
\usepackage{graphicx}
\usepackage{subcaption}
\usepackage{algorithm}
\usepackage{algorithmic}
\usepackage{multirow}
\usepackage{enumitem}
\usepackage{wrapfig}
\usepackage{tikz}
\usetikzlibrary{shapes, arrows, positioning, fit, backgrounds}

\usepackage{pifont}
\newcommand{\cmark}{\ding{51}}
\newcommand{\xmark}{\ding{55}}

\title{CosFly-Track: A Large-Scale Multi-Modal Dataset for UAV Visual Tracking via Multi-Constraint Trajectory Optimization}

\author{
  Xiangyue Wang$^{1}$ \quad
  Hanxuan Chen$^{1}$ \quad
  Songsheng Cheng$^{1}$\thanks{Equal contribution.} \quad
  Ruilong Ren$^{1}$\footnotemark[1] \quad
  Jie Zheng$^{2}$\footnotemark[1] \\
 \textbf{ Shuai Yuan$^{3}$\footnotemark[1] \quad
  Tianle Zeng$^{4}$ \quad
  Hanzhong Guo$^{5}$ \quad
  Kangli Wang$^{1}$\thanks{Corresponding authors.} \quad
  Ji Pei$^{1}$\footnotemark[2] }\\
  $^{1}$Autel Robotics \quad
  $^{2}$Nanjing University \quad
  $^{3}$Peking University \\
  $^{4}$Southern University of Science and Technology \quad
  $^{5}$University of Hong Kong \\
  \texttt{peiji@autelrobotics.com}
}

\begin{document}

\maketitle

\begin{abstract}
Recent aerial vision-language navigation (VLN) datasets have grown rapidly, but they primarily address goal-oriented navigation to static destinations, leaving UAV visual tracking---continuously following a moving target while maintaining visibility---largely without dedicated training data.
We introduce \textbf{CosFly-Track}, a large-scale multi-modal dataset and scalable generation pipeline for UAV visual tracking in urban environments.
The dataset provides approximately 12{,}000 expert and perturbed UAV trajectories generated from 6{,}000 pedestrian paths, comprising 2.4 million timesteps ($\sim$334 hours) with 7 aligned data channels: RGB, metric depth, semantic segmentation, 6-DoF drone pose, target state with visibility flag, bilingual (Chinese--English) instructions, and trajectory-pair metadata.
To generate high-quality expert trajectories, we develop \textbf{MuCO}, a multi-constraint optimizer that plans directly in continuous 3D space with BVH-accelerated collision and visibility queries, jointly enforcing target visibility, viewpoint quality, collision avoidance, smoothness, and kinematic feasibility---avoiding the discretization artifacts and post-hoc smoothing of grid-based planners.
Fine-tuning experiments on seven vision-language models show that CosFly-Track improves tracking performance to 78.3--95.6\% SR@1m---a 53--69 percentage-point gain over zero-shot baselines---supporting the dataset as a training resource for dynamic target-following agents.
The dataset is publicly available at \url{https://huggingface.co/datasets/AutelRobotics/CosFly}; evaluation scripts and pre-trained checkpoints are hosted at \url{https://huggingface.co/AutelRobotics/CosFly-Track}.
\end{abstract}

\section{Introduction}
\label{sec:intro}

Aerial vision-language navigation (VLN) datasets have grown rapidly~\citep{openfly2026, airnav2026, citynav2025, aerialvln2023, indooruav2026}, reflecting increasing interest in enabling autonomous UAVs to understand and execute natural-language instructions in complex environments.
However, existing aerial VLN datasets are predominantly designed for \textbf{goal-oriented navigation}---planning a path from a start location to a fixed destination.
To our knowledge, no existing dataset is designed for the substantially different task of \emph{visual tracking}: continuously following a \emph{dynamic} target while maintaining target visibility, avoiding collisions, and satisfying kinematic constraints.

\begin{figure}[t]
\centering
\includegraphics[width=\linewidth]{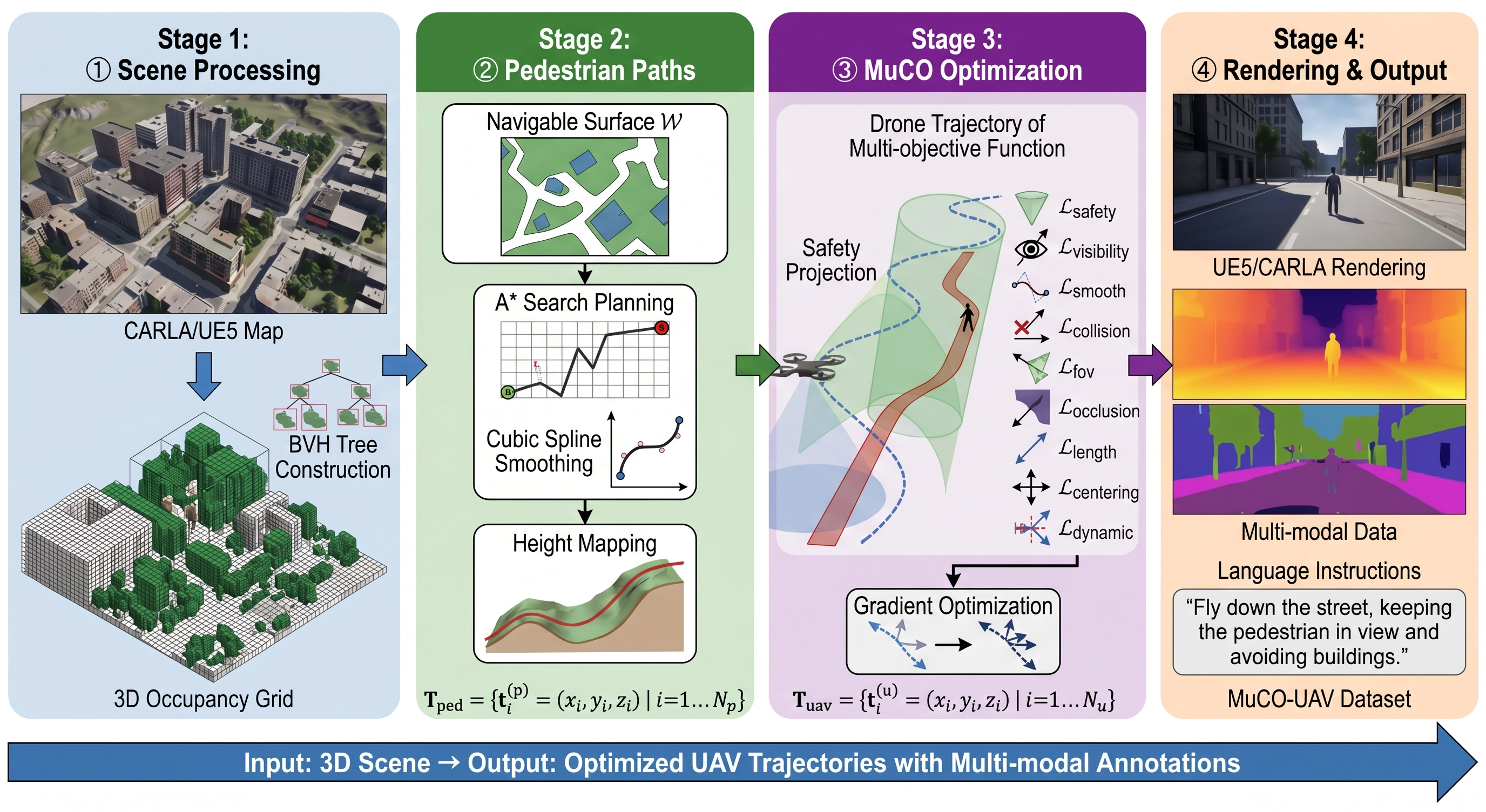}
\caption{
    \textbf{CosFly-Track pipeline.}
    From urban scenes to dataset: 3D grid construction $\to$ pedestrian path generation $\to$ MuCO trajectory optimization (9-term objective, soft/hard constraints) $\to$ paired expert/perturbed rendering with 7 aligned data channels. Details in Section~\ref{sec:method}.
  }
  \label{fig:teaser}
\end{figure}

Navigation aims to reach a fixed goal; tracking requires \emph{continuously} adapting to a moving target under visibility, viewpoint, collision, and kinematic constraints at every timestep (Figure~\ref{fig:teaser}; formal definitions in Section~\ref{sec:task}).
This gap matters because UAV visual tracking underlies applications such as search and rescue, autonomous cinematography, sports analysis, wildlife monitoring, and infrastructure inspection.
These scenarios require not only identifying a target, but also generating executable UAV motion that preserves target visibility over extended periods.
Yet there is still no dedicated large-scale multi-modal dataset for training and evaluating agents under these tracking-specific constraints.

\paragraph{Challenges of tracking data generation.}
Generating high-quality tracking trajectories poses additional challenges beyond navigation path generation:
(1)~Grid-based planners such as A$^*$ optimize geometric path length and often require post-processing to satisfy UAV velocity, acceleration, and jerk constraints;
(2)~Tracking trajectories must jointly optimize target visibility, viewpoint quality, distance control, and collision avoidance as the target moves---objectives absent from navigation planners;
(3)~Scaling such planning to dense urban scenes is computationally demanding because the search space grows rapidly with map resolution and trajectory length.
These challenges motivate a continuous multi-constraint optimization approach tailored to UAV tracking.

\paragraph{Contributions.}
We make the following contributions:
\begin{enumerate}[leftmargin=*,nosep]
\item \textbf{Dataset}: To our knowledge, CosFly-Track is the first large-scale multi-modal dataset for UAV visual tracking, providing $\sim$12K expert and perturbed trajectories generated from $\sim$6K pedestrian paths, 2.4M timesteps, 7 aligned data channels, and bilingual instructions (Section~\ref{sec:dataset}).
\item \textbf{Generation pipeline}: We develop CosFly, a modular data production pipeline built around MuCO, a multi-constraint trajectory optimizer that jointly considers target visibility, viewpoint quality, obstacle avoidance, and kinematic feasibility in continuous 3D space (Section~\ref{sec:method}).
\item \textbf{Benchmarks}: We evaluate seven VLMs on the proposed tracking task, showing that fine-tuning on CosFly-Track yields a 53--69 percentage-point improvement in SR@1m over zero-shot baselines, along with scaling analysis and ablation studies (Section~\ref{sec:experiments}).
\item \textbf{Open resources}: We release the dataset, evaluation scripts, and pre-trained checkpoints to support future research on UAV visual tracking.
\end{enumerate}

\section{Related Work}
\label{sec:related}

\subsection{UAV Visual Datasets}

\paragraph{Real-world UAV datasets.}
Early UAV datasets focus on object detection and tracking from fixed or pre-programmed trajectories: VisDrone~\citep{visdrone2018} provides bounding-box annotations for detection/tracking; UAV123~\citep{uav1232016} benchmarks single-object tracking; UAVDT~\citep{uavdt2018} targets detection under adverse conditions.
These datasets lack action-level annotations (drone control commands) and language instructions, making them unsuitable for training autonomous tracking agents.

\paragraph{Simulated aerial VLN datasets.}
The recent surge in aerial VLN~\citep{anderson2018evaluation, vlnce2020} has produced several navigation-focused datasets:
AerialVLN~\citep{aerialvln2023} provides 8.4K trajectories in AirSim~\citep{airsim2018} for navigation;
CityNav~\citep{citynav2025} scales to 32.6K trajectories in real-world aerial imagery;
OpenFly~\citep{openfly2026} offers 100K trajectories across 18 scenes using A$^*$ voxel planning;
AirNav~\citep{airnav2026} provides 143K trajectories from SfM-reconstructed environments;
IndoorUAV~\citep{indooruav2026} targets indoor navigation with 16K+ trajectories.
These datasets all target navigation to static goals; no existing dataset addresses the dynamic-target tracking task that CosFly-Track is designed for.

\subsection{Trajectory Planning and Optimization}

Discrete planners such as A$^*$~\citep{astar1968}, D$^*$ Lite~\citep{dstar2002}, and RRT~\citep{rrt1998} operate on grid or sampling-based representations, often producing paths that are kinematically infeasible and require post-processing smoothing.
A common remedy is to apply B-spline or polynomial smoothing after discrete search, but the smoothed trajectory still inherits the suboptimal topology of the grid search and cannot jointly optimize visibility and kinematics end-to-end.
Continuous optimization methods---CHOMP~\citep{chomp2009}, TrajOpt~\citep{trajopt2014}, GPMP~\citep{gpmp2016}---optimize trajectories in $\mathbb{R}^3$ using gradient-based methods.
MuCO differs from these approaches in its \emph{tracking-specific} cost design: it jointly optimizes 9 objectives, including BVH-accelerated visibility checking, direction-aware viewpoint cost, and jerk regularization, with a novel soft/hard constraint architecture for safety guarantees.
We empirically show that MuCO achieves quality comparable to A$^*$~\citep{astar1968} while being $22\times$ faster (Section~\ref{sec:method}).

\subsection{VLMs in Embodied AI}

Vision-language models~\citep{llava2023, qwenvl2023, internvl2024} have demonstrated strong capabilities in visual understanding and instruction following.
Recent work~\citep{rt22023, embodiedbench2025} explores fine-tuning VLMs for embodied control tasks.
To our knowledge, CosFly-Track provides the first large-scale training resource for adapting VLMs to the UAV tracking task, with bilingual instructions enabling cross-lingual research.

\section{Task Formulation}
\label{sec:task}

\subsection{Tracking vs.\ Navigation}

\begin{table}[t]
\caption{
  \textbf{Comparison with existing aerial VLN datasets.}
  CosFly-Track is the \emph{only} dataset targeting the tracking task, offering 6-DoF actions, kinematic constraints, visibility guarantees, paired trajectories, and 7 aligned data channels (RGB, depth, segmentation, 6-DoF pose, target state, bilingual instructions, trajectory-pair metadata).
  Avg.~Len = average trajectory length in steps.
  $^\dagger$Initial release; expansion to 100K+ trajectories ($\sim$20M frames) is underway.
}
\label{tab:comparison}
\centering
\small
\setlength{\tabcolsep}{3.5pt}
\begin{tabular}{l c c r r c c c c c}
\toprule
Dataset & Venue & Task & Traj. & Avg.~Len & DoF & Mod. & Kinem. & Vis. & Paired \\
\midrule
VisDrone~\citep{visdrone2018} & arXiv'18 & Det. & --- & --- & --- & RGB & --- & \xmark & \xmark \\
UAV123~\citep{uav1232016} & ECCV'16 & SOT & 123 & --- & --- & RGB & --- & \xmark & \xmark \\
AVDN~\citep{avdn2023} & ACL'23 & \textit{Nav.} & 3K & 49 & 3 & RGB & \xmark & \xmark & \xmark \\
AerialVLN~\citep{aerialvln2023} & ICCV'23 & \textit{Nav.} & 8.4K & 230 & 4 & RGB & \xmark & \xmark & \xmark \\
OpenUAV~\citep{openuav2024} & ICLR'25 & \textit{Nav.} & 12.1K & 264 & 6 & RGB & \xmark & \xmark & \xmark \\
CityNav~\citep{citynav2025} & ICCV'25 & \textit{Nav.} & 32.6K & 240 & 4 & RGB & --- & \xmark & \xmark \\
IndoorUAV~\citep{indooruav2026} & AAAI'26 & \textit{Nav.} & 16K+ & 22 & 4 & RGB+D & \xmark & \xmark & \xmark \\
OpenFly~\citep{openfly2026} & ICLR'26 & \textit{Nav.} & 100K & 35 & 4 & RGB & \xmark & \xmark & \xmark \\
AirNav~\citep{airnav2026} & arXiv'26 & \textit{Nav.} & 143K & 193 & 4 & RGB & \xmark & \xmark & \xmark \\
\midrule
\textbf{CosFly-Track} & arXiv'26 & \textbf{Trk.} & \textbf{12K}$^\dagger$ & \textbf{200} & \textbf{6} & \textbf{7} & \cmark & \cmark & \cmark \\
\bottomrule
\end{tabular}
\end{table}

We formally distinguish the UAV visual tracking task from navigation (Table~\ref{tab:comparison}):

\textbf{Navigation.} Given a static goal position $g$ (or language instruction), plan a path from start $s$ to $g$ in a static environment. Success is measured by reaching the goal.

\textbf{Visual tracking.} Given a dynamic target with trajectory $\{p_1, \ldots, p_T\}$, continuously follow the target while maintaining visibility and satisfying kinematic constraints. Success requires \emph{sustained} tracking---the target must remain visible and within appropriate range throughout the entire trajectory.

\subsection{Input and Action Spaces}

The tracking agent receives:
\textbf{RGB image} $o_t \in \mathbb{R}^{H \times W \times 3}$ (optionally with depth and segmentation);
\textbf{6-DoF pose history} $\{(x, y, z, \phi, \theta, \psi)\}_{t-k}^{t}$;
\textbf{target bounding box and visibility flag};
\textbf{bilingual tracking instruction}.
The action space consists of 6-DoF incremental waypoint predictions $a_t = (\Delta x, \Delta y, \Delta z, \Delta \phi, \Delta \theta, \Delta \psi)$.

\subsection{Evaluation Metrics}

\paragraph{Waypoint prediction.}
\textbf{SR@$r$}: percentage of final waypoints within $r$\,m of ground truth ($r \in \{0.5, 1.0, 2.0\}$);
\textbf{ADE}/\textbf{FDE}: average/final displacement error (m);
\textbf{RotAcc@$d$}: percentage of final waypoints within $d^\circ$ of ground truth yaw;
\textbf{Yaw MAE}: mean absolute yaw error (degrees);
\textbf{JointSR@($r$,$d$)}: joint position-rotation success;
\textbf{MAE}: mean absolute error across all 6 DoF.

\paragraph{Target bbox prediction.}
\textbf{mIoU}: mean intersection-over-union between predicted and ground-truth bounding boxes.

\paragraph{Metric usage.}
The architecture comparison uses SR@1m, FDE, RotAcc@$1^\circ$, MAE for coarse-to-fine ranking; the scaling analysis reports SR@2m and IoU$\geq$0.75 on the hard split where SR@1m saturates; ablation studies report the full metric set (ADE, FDE, SR@1m, Yaw MAE, mIoU) for comprehensive modality and paradigm analysis.

\section{The CosFly Pipeline and MuCO Engine}
\label{sec:method}

\subsection{CosFly Pipeline Overview}

The CosFly pipeline is a modular data production system for aerial tracking datasets, described in full engineering detail in our technical report~\citep{cosfly_report2026}, which documents the complete engineering implementation of this work.
Here we summarize the six stages:
\textbf{(1)~Environment preprocessing}---extract and simplify 3D AABB obstacle maps (65K$\to$2K boxes);
\textbf{(2)~Pedestrian trajectory generation}---A$^*$ on walkable grids with variable-speed resampling;
\textbf{(3)~MuCO tracking optimization}---detailed in \S\ref{sec:muco};
\textbf{(4)~Dual-trajectory augmentation}---paired expert-perturbed trajectories via Bernoulli perturbations;
\textbf{(5)~Multi-modal rendering}---RGB, depth, segmentation, 6-DoF pose, target state with weather randomization;
\textbf{(6)~Bilingual captioning}---teacher-student distillation (Qwen3.5-397B$\to$2B~\citep{qwen3.5}) for Chinese/English tracking instructions.
The pipeline is fully decoupled: each stage can be independently replaced (e.g., substituting real-world GPS traces for simulated pedestrian trajectories, or a different rendering engine).
This paper focuses on Stages 3--4 (MuCO optimizer and dual-trajectory design) and the downstream benchmarking experiments; readers are referred to our technical report~\citep{cosfly_report2026} for complete details of map processing, quality inspection, zoom simulation, and caption generation.

\subsection{MuCO: Multi-Constraint Trajectory Optimizer}
\label{sec:muco}

\paragraph{Problem formulation.}
Given an $N$-step pedestrian trajectory $\{p_1, \ldots, p_N\}$ and obstacle map $\mathcal{O}$, MuCO optimizes waypoints $\mathbf{W} = \{w_1, \ldots, w_N\}$, $w_i \in \mathbb{R}^3$, by minimizing:
\begin{equation}
C(\mathbf{W}) = \sum_{i=1}^{N} \sum_{k=1}^{9} \lambda_k \cdot c_k(w_i, \text{context}_i)\,,
\label{eq:cost}
\end{equation}
where $c_k$ denotes one of the 9 cost terms and $\lambda_k$ denotes its weight.

\paragraph{Cost function design.}
We highlight three key cost terms (all 9 in Appendix~\ref{app:muco}).
\textbf{Visibility}: $c_{\text{vis}}(w_i) = (1 - v_i)^2$, where $v_i$ is the fraction of unoccluded ray samples (5--10 points on the target body); ray-obstacle queries are accelerated from $O(n)$ to $O(\log n)$ via BVH.
\textbf{Viewpoint}: $c_{\text{view}}(w_i) = f(\cos\langle -\vec{v}_{\text{look}}, \vec{v}_{\text{walk}}\rangle) \cdot d_i$, where $\vec{v}_{\text{walk}}$ uses net displacement over a $\pm$15-frame window and the directness factor $d_i$ automatically reduces the cost during target turns.
\textbf{Jerk}: $c_{\text{jerk}}(w_i) = \|w_{i+2} - 3w_{i+1} + 3w_i - w_{i-1}\|^2$, a third-order difference penalty ensuring kinematic executability.
The remaining costs cover tracking distance, smoothness, safety, pitch angle, altitude, and path length.

\paragraph{Soft/hard constraint architecture.}
Safety is guaranteed through four layers:
(1)~\emph{soft safety cost} provides gradient guidance away from obstacles;
(2)~\emph{geometric projection} iteratively pushes waypoints out of obstacles via vertical lift, horizontal bypass, or local displacement;
(3)~\emph{velocity repair} redistributes projection-induced spikes to subsequent waypoints;
(4)~\emph{altitude smoothing} eliminates oscillations.
This decoupled design prevents the optimizer from being trapped by overly aggressive safety penalties while guaranteeing collision-free trajectories.

\paragraph{Optimization.}
We use coordinate-wise finite differences with $\varepsilon = 0.5$\,m---large enough to cross small obstacles (e.g., tree canopies) for effective visibility gradients, with adaptive learning rate (halving on cost increase, slowly recovering on decrease) and per-waypoint displacement clipping of 0.5\,m. Rayon-based parallel gradient computation achieves 2--5\,ms per iteration for 200 waypoints.

\subsection{Performance}

BVH acceleration reduces per-trajectory optimization from $\sim$12\,s to 0.1--0.5\,s (24--120$\times$ speedup; detailed in Table~\ref{tab:performance} in the appendix). Per-path obstacle filtering further reduces BVH size from $\sim$2{,}000 to $\sim$200 obstacles.

\paragraph{MuCO vs.\ A$^*$~\citep{astar1968}.}
We implement an A$^*$-based tracking planner that searches over a 4D spatiotemporal voxel graph (position $\times$ timestep) with BVH-accelerated visibility costs and safety-distance hard constraints---the strongest discrete-planning baseline. Table~\ref{tab:muco_vs_astar} compares both planners on 20 diverse trajectories from the same environment.

\begin{table}[t]
\caption{
  \textbf{MuCO vs.\ A$^*$} on 20 shared pedestrian trajectories.
  MuCO achieves comparable visibility and tracking distance while being $22\times$ faster with 13\% shorter paths (Figure~\ref{fig:muco_vs_astar}).
}
\label{tab:muco_vs_astar}
\centering
\small
\begin{tabular}{l c c c c c}
\toprule
Planner & Visibility$\uparrow$ & Track Dist.~(m) & Path Len.~(m)$\downarrow$ & Time~(ms)$\downarrow$ & Nodes \\
\midrule
A$^*$   & \textbf{0.979} & 28.20 & 125.5 & 5{,}467 & 400K \\
MuCO    & 0.906          & 27.96 & \textbf{108.8} & \textbf{247} & --- \\
\midrule
$\Delta$ & $-$7.3\% & $-$0.24 & $-$13.3\% & $\mathbf{22\times}$ & --- \\
\bottomrule
\end{tabular}
\end{table}

\begin{figure}[t]
\centering
\includegraphics[width=\linewidth]{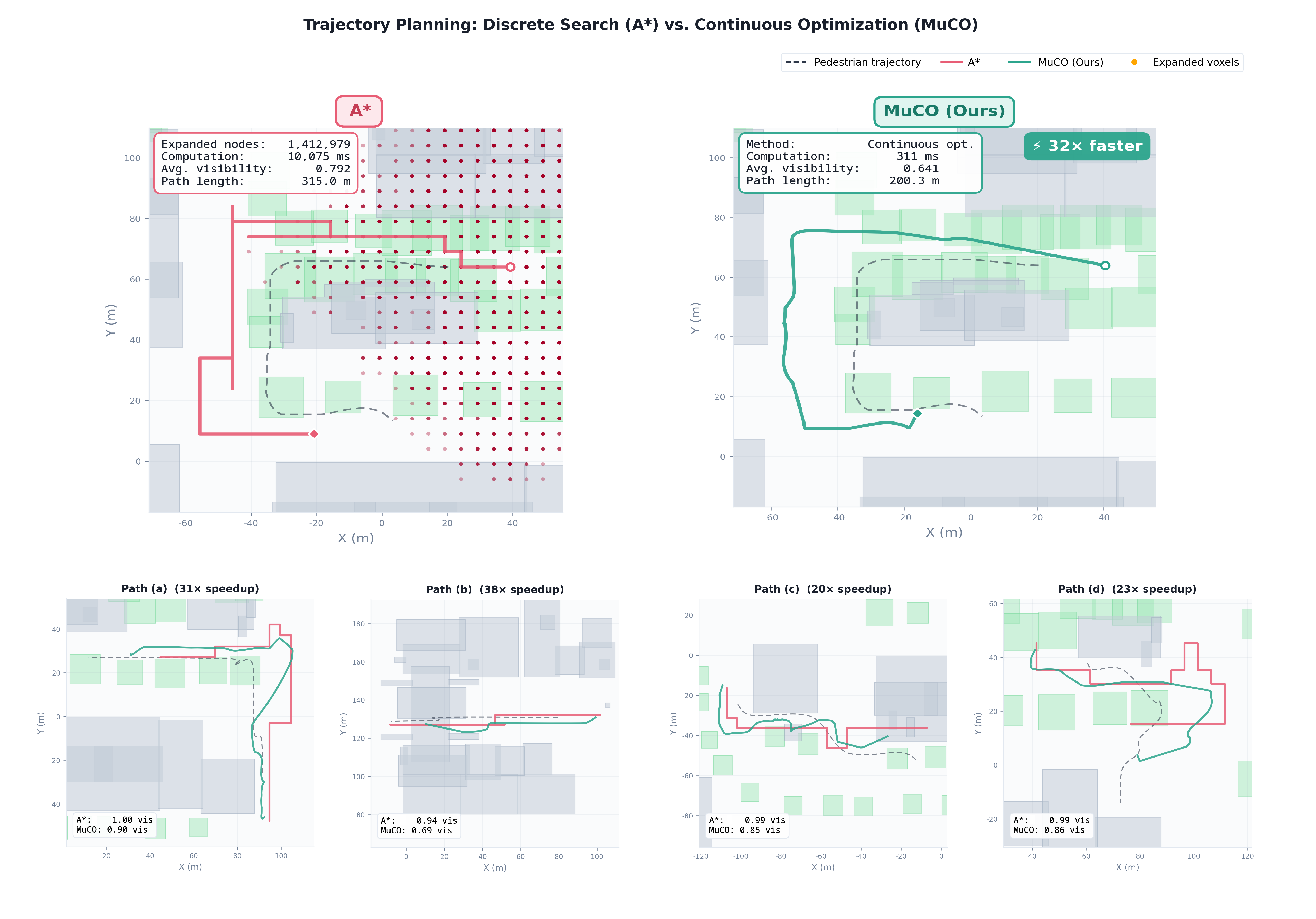}
\caption{
  \textbf{MuCO vs.\ A$^*$.}
  \emph{Left}: A$^*$ expands 1.4M voxels on a 315\,m path (10s, visibility 0.79); MuCO produces a smooth trajectory in 311\,ms (32$\times$ faster, visibility 0.64).
  \emph{Right}: four additional scenarios showing 20--32$\times$ speedup with comparable tracking quality. Dashed lines: pedestrian paths.
}
\label{fig:muco_vs_astar}
\end{figure}

As shown in Figure~\ref{fig:muco_vs_astar}, A$^*$ achieves +7.3\% higher visibility by searching a dense voxel space ($\sim$400K node expansions per trajectory), but this comes at $22\times$ higher latency (5.5\,s vs.\ 247\,ms).
In practice, the visibility gap is concentrated in a few hard trajectories with severe occlusion (e.g., tight alleys); on 16 of 20 trajectories MuCO maintains visibility $>$0.90.
Crucially, MuCO produces 13\% shorter paths with better kinematic smoothness (jerk-regularized continuous optimization), resulting in more natural drone motion.
At the dataset-generation scale ($>$6{,}000 trajectories), A$^*$ would require $\sim$9 GPU-hours, while MuCO completes in $\sim$25 minutes---making continuous optimization a practical approach for large-scale production.

\section{The CosFly-Track Dataset}
\label{sec:dataset}

\subsection{Scale and Statistics}

CosFly-Track provides the data summarized in Table~\ref{tab:dataset_stats}.

\begin{table}[t]
\caption{
  \textbf{CosFly-Track dataset statistics.}
  Each pedestrian path generates an expert (MuCO-optimized) and a perturbed trajectory, yielding paired data for diverse training paradigms.
}
\label{tab:dataset_stats}
\centering
\small
\begin{tabular}{l r}
\toprule
Statistic & Value \\
\midrule
Unique pedestrian paths & $\sim$6{,}000 \\
Expert/perturbed trajectories & $\sim$12{,}000 \\
Timesteps (expert + perturbed) & 2.4M \\
Annotation samples (2.4M $\times$ modalities) & $\sim$10M \\
Total tracking duration & $\sim$334 hours \\
Urban scenes (CARLA~\citep{carla2017} Towns) & 16 \\
Weather/lighting conditions & 8+ \\
Average trajectory length & $\sim$200 steps / 100\,s \\
\bottomrule
\end{tabular}
\end{table}

Each trajectory sample combines 7 aligned data channels. Five are recorded at every timestep:
(1)~\textbf{RGB image} (1280$\times$720);
(2)~\textbf{metric depth} (float32, per-pixel meters);
(3)~\textbf{semantic segmentation} (CARLA labels);
(4)~\textbf{6-DoF drone pose} $(x,y,z,\phi,\theta,\psi)$;
(5)~\textbf{target state} (world coordinates + visibility flag);
Two additional annotations are provided at the trajectory level:
(6)~\textbf{bilingual tracking instructions} (Chinese/English);
(7)~\textbf{trajectory-pair metadata} (expert/perturbed label + temporal alignment index).

\subsection{Dual-Trajectory Augmentation}

Each pedestrian path generates \emph{two} UAV trajectories: an \textbf{expert trajectory} (MuCO-optimized) and a \textbf{perturbed trajectory} that simulates realistic tracking errors via a joint perturbation framework.
At each frame, two independent Bernoulli events control position perturbation ($P_{\text{pos}}=0.6$, displacement radius 2--3\,m for pedestrian/drone) and rotation perturbation ($P_{\text{rot}}=0.6$, max viewing-angle deviation 5$^\circ$), yielding four augmentation states: 16\% unperturbed, 24\% position-only, 24\% rotation-only, and 36\% fully perturbed.
A sliding-window sampler (window size 10, stride 3) constructs multi-task training samples: the first 5 frames use \emph{perturbed} observations as input, while the full 10-frame \emph{expert} trajectory provides ground-truth supervision for both denoising (frames 0--4) and prediction (frames 5--9).
This paired design supports diverse training paradigms:
(1)~\emph{denoising}---recover expert actions from perturbed observations;
(2)~\emph{DAgger-style correction}~\citep{ross2011dagger}---train on mixed expert/perturbed data to handle distribution shift;
(3)~\emph{contrastive learning}---distinguish high- from low-quality tracking;
(4)~\emph{prediction}---forecast future expert waypoints from noisy history.

\subsection{Additional Features and Reproducibility}

Each trajectory includes bilingual (Chinese/English) natural language instructions (e.g., ``Follow the pedestrian walking north, maintaining 25\,m distance and 45$^\circ$ pitch''), enabling VLM instruction fine-tuning and cross-lingual research.
The CosFly pipeline has been validated for migration from CARLA (UE4) to SimWorld (UE5); the modular design supports replacing the optimizer with task-specific objectives (e.g., inspection, patrol).

An initial subset ($\sim$100K multi-modal frames) is publicly available at \url{https://huggingface.co/datasets/AutelRobotics/CosFly}, with progressive expansion toward the full $\sim$2M frames planned.
Evaluation scripts and pre-trained checkpoints are hosted at \url{https://huggingface.co/AutelRobotics/CosFly-Track}.
We additionally provide Croissant metadata~\citep{croissant2024} with RAI fields.
Complete pipeline engineering details are documented in our technical report~\citep{cosfly_report2026}.

\paragraph{Data quality.}
We retain a small proportion ($<$5\%) of imperfect samples (e.g., target--vehicle overlap, background pedestrian frame-skipping) rather than aggressively filtering them; empirical evaluation confirms negligible impact on model performance while preserving natural scene diversity.

\section{Experiments and Benchmarks}
\label{sec:experiments}

\subsection{Experimental Setup}

\paragraph{Models.}
We evaluate seven widely used VLMs spanning 0.8B--9B parameters:
Qwen3.5-0.8B, Qwen3.5-2B, Qwen3.5-9B~\citep{qwen3.5}, Qwen3-VL-2B, Qwen3-VL-8B~\citep{qwen3technicalreport}, GLM-4.6V-Flash~\citep{glm46v}, and Gemma-4-E4B~\citep{gemma4}.

\paragraph{Task and training protocol.}
Given four history frames plus the current frame (five RGB images total), together with the drone's current 6-DoF pose and target bounding box, the model predicts the next five waypoints as 6-DoF increments $(\Delta x, \Delta y, \Delta z, \Delta\text{pitch}, \Delta\text{yaw}, \Delta\text{roll})$.

We use three experimental configurations drawn from different subsets of the full 2.4M-timestep dataset:
\begin{itemize}[leftmargin=*,nosep]
\item \textbf{Architecture comparison} (\S\ref{sec:experiments}.2): Full-parameter SFT (vision encoder frozen), 200K samples from 16 CARLA maps, 1 epoch, batch size 64, learning rate $5{\times}10^{-6}$, evaluation on 1{,}160 held-out samples.
\item \textbf{Scaling analysis} (\S\ref{sec:experiments}.3): LoRA~\citep{hu2022lora} fine-tuning (r=64, $\alpha$=128), 250K--1M samples, evaluation on hard-difficulty split.
\item \textbf{Ablation studies} (\S\ref{sec:experiments}.4): LoRA fine-tuning, 213K samples (760 trajectories), 1{,}672 fixed steps, evaluation on 11{,}878 samples stratified by difficulty and scene familiarity.
\end{itemize}
All configurations use DeepSpeed ZeRO-2~\citep{rajbhandari2020zero} and bf16 precision.

\paragraph{Evaluation metrics.}
SR@1m, SR@0.5m, RotAcc@$1^\circ$, ADE, FDE, MAE, JointSR@(0.5\,m,$1^\circ$).

\subsection{Main Results: VLM Architecture Comparison}

\begin{table}[t]
\caption{
  \textbf{VLM architecture comparison.}
  SFT fine-tuning on 200K CosFly-Track samples brings a 53--69 percentage-point improvement in SR@1m over zero-shot.
  $\ddagger$Zero-shot outputs match the \emph{predict-zero} baseline (all deltas = 0) to 4 decimal places, confirming models are not directly usable without task-specific fine-tuning.
}
\label{tab:main_results}
\centering
\small
\begin{tabular}{l c c c c c}
\toprule
Model & Params & SR@1m$\uparrow$ & FDE$\downarrow$ & RotAcc@$1^\circ$$\uparrow$ & MAE$\downarrow$ \\
\midrule
\multicolumn{6}{c}{\textit{Zero-shot (no fine-tuning)$^\ddagger$}} \\
\midrule
Qwen3.5-0.8B~\citep{qwen3.5} & 0.8B & 25.17 & 2.872 & 40.41 & 1.731 \\
Qwen3.5-2B & 2B & 25.17 & 2.872 & 40.41 & 1.731 \\
Qwen3-VL-2B~\citep{qwen3technicalreport} & 2B & 25.17 & 2.872 & 40.41 & 1.731 \\
Qwen3-VL-8B~\citep{qwen3technicalreport} & 8B & 33.47 & 2.601 & 40.78 & 1.527 \\
\midrule
\multicolumn{6}{c}{\textit{SFT fine-tuned on CosFly-Track}} \\
\midrule
Gemma-4-E4B~\citep{gemma4} & 4B & 78.34 & 1.527 & 72.33 & 0.461 \\
Qwen3.5-0.8B & 0.8B & 94.50 & 1.298 & 87.61 & 0.334 \\
Qwen3.5-2B & 2B & 95.08 & 1.271 & 93.12 & 0.298 \\
Qwen3-VL-2B & 2B & 94.80 & 1.315 & 88.45 & 0.352 \\
Qwen3-VL-8B & 8B & 95.22 & 1.247 & 91.88 & 0.312 \\
Qwen3.5-9B~\citep{qwen3.5} & 9B & \textbf{95.60} & \textbf{1.213} & \textbf{95.86} & \textbf{0.268} \\
GLM-4.6V-Flash~\citep{glm46v} & 4.6B & 95.48 & 1.231 & 95.77 & 0.274 \\
\bottomrule
\end{tabular}
\end{table}

\paragraph{Key findings.}
Table~\ref{tab:main_results} summarizes the results.
(1)~Zero-shot VLMs are essentially non-functional for tracking: all models match the \emph{predict-zero} baseline (SR@1m 25--33\%), confirming that this task requires task-specific SFT.
(2)~SFT brings substantial improvements across all models (a 53--69 percentage-point gain in SR@1m and a 16--31 percentage-point gain in RotAcc@$1^\circ$), supporting CosFly-Track's utility as training data.
(3)~Within the Qwen3.5 family (0.8B$\to$2B$\to$9B), SR@1m improves by only 0.5 percentage points, but RotAcc@$1^\circ$ jumps by 7.0 percentage points and MAE drops by 18.3\%---model capacity mainly benefits \emph{fine-grained} rotation prediction rather than coarse position accuracy.
(4)~The Qwen3-VL series (2B/8B) underperforms similarly-sized Qwen3.5 models after SFT (MAE nearly $2\times$ worse for the 8B variant), suggesting that its visual encoder is less suited for geometric regression.
(5)~Gemma-4-E4B significantly underperforms (SR@1m 78.34\%, loss plateaued at 0.73), likely because its sliding-window attention is less well matched to sequential waypoint regression.

\subsection{Data Scaling Analysis}

We train Qwen3.5-2B and Qwen3.5-0.8B (LoRA) on 25\%/50\%/100\% of the data (250K--1M samples) and evaluate on the hard-difficulty split (full results in Table~\ref{tab:scaling}, appendix).
Both models improve monotonically: ADE decreases $\sim$2\% and SR@2m increases $\sim$1.5 percentage points from 25\% to 100\%, but gains \emph{do not saturate}, supporting further scaling.
The 2B and 0.8B models converge to nearly identical ADE ($\sim$2.14) and FDE ($\sim$2.64), suggesting the current data distribution---rather than model capacity---is the primary constraint.

\subsection{Data Composition Ablation}

We conduct comprehensive ablations on input modalities and training data composition using Qwen3.5-0.8B with LoRA (r=64, $\alpha$=128) fine-tuning. All seven configurations share identical hyperparameters (learning rate $10^{-5}$, cosine schedule, global batch 128, max 1{,}672 steps) and the same 213K training / 11{,}878 evaluation split.

\begin{table}[t]
\caption{
  \textbf{Data composition ablation} (Qwen3.5-0.8B~\citep{qwen3.5} LoRA, 213K samples).
  \emph{Top}: input modality ablation---removing pose causes $3.1\times$ FDE increase.
  \emph{Bottom}: training paradigm ablation---denoising achieves the best FDE/SR@1m while expert-only degrades yaw ($1.7\times$).
}
\label{tab:ablation}
\centering
\small
\setlength{\tabcolsep}{3.5pt}
\begin{tabular}{l c c c c c c}
\toprule
Configuration & ADE$\downarrow$ & FDE$\downarrow$ & SR@1m$\uparrow$ & Yaw$\downarrow$ & mIoU$\uparrow$ \\
\midrule
\multicolumn{6}{c}{\textit{Input modality}} \\
\midrule
Pose+BBox (no RGB)      & 0.849 & 1.264 & 77.1 & 3.85$^\circ$ & 0.605 \\
RGB+Pose                & 0.843 & 1.248 & 77.5 & 4.11$^\circ$ & 0.479 \\
RGB+BBox                & 2.439 & 3.805 & 17.6 & 3.88$^\circ$ & 0.599 \\
RGB only                & 2.503 & 3.911 & 15.8 & 3.93$^\circ$ & 0.487 \\
RGB+Pose+BBox (full)    & 0.846 & 1.249 & 77.6 & 3.87$^\circ$ & 0.603 \\
\midrule
\multicolumn{6}{c}{\textit{Training paradigm (full input)}} \\
\midrule
Expert + Perturbed (mixed)                    & 0.846 & 1.249 & 77.6 & 3.87$^\circ$ & 0.603 \\
Expert-only                                   & 0.855 & 1.270 & 75.1 & 6.54$^\circ$ & 0.560 \\
\textbf{Perturbed$\to$Expert (denoising)}     & \textbf{0.844} & \textbf{1.239} & \textbf{78.8} & 3.97$^\circ$ & \textbf{0.602} \\
\bottomrule
\end{tabular}
\end{table}

\paragraph{Modality ablation} (Table~\ref{tab:ablation}, top).
(1)~\textbf{Pose history is the decisive input}: removing it causes FDE to increase $3.1\times$ (1.25$\to$3.85\,m) and SR@1m to collapse from 77.6\% to 15--17\%.
All five configurations with pose cluster in FDE 1.24--1.27\,m regardless of other modalities, while both no-pose configurations degrade to FDE $>$3.8\,m even on easy trajectories.
(2)~BBox history is critical for target prediction: mIoU drops from 0.60 to 0.48 ($-20\%$) and IoU$\geq$0.75 drops from 0.56 to 0.31 ($-45\%$) without it.
(3)~When pose + bbox are both present, RGB provides marginal benefit (FDE 1.264 vs.\ 1.249), suggesting the current benchmark's structured text priors suffice for path regression. This does \emph{not} imply RGB is generally unnecessary---more challenging out-of-distribution scenarios (sharp turns, intent changes) may require visual cues.

\paragraph{Training paradigm ablation} (Table~\ref{tab:ablation}, bottom).
The denoising paradigm (perturbed input $\to$ expert target) achieves the best overall FDE (1.239) and SR@1m (78.8\%), outperforming both expert-only and mixed training.
Expert-only training severely degrades yaw prediction (MAE 6.54$^\circ$ vs.\ 3.85$^\circ$), because models overfit to the narrow distribution of clean expert trajectories and fail to learn corrective heading adjustments.
This validates the dual-trajectory design as a core contribution: perturbation-augmented training is essential for learning robust orientation control.

\subsection{Cross-Scene Generalization}

We compare single-map vs.\ multi-map training (Qwen3.5-0.8B, 200K samples, 8 held-out test maps; Table~\ref{tab:cross_scene} in the appendix).
While SR@1m saturates above 95\% for both settings (+0.11 percentage points), strict metrics reveal substantial structural gains: JointSR@(0.5\,m,$1^\circ$) improves by +5.31 percentage points, rotation MAE drops 12.5\%, and catastrophic failures (FDE$>$10\,m) decrease by 55.6\%.
The improvement holds on the fully out-of-distribution Town10HD (+5.20 percentage points in JointSR), suggesting that multi-map training learns generalizable geometric awareness.

\subsection{Downstream Task Transfer}

Beyond tracking, CosFly-Track's multi-modal annotations support depth estimation, instance segmentation, and object detection (Table~\ref{tab:downstream} in the appendix).
Fine-tuning on $\sim$100K frames substantially improves all tasks: Depth Anything V2~\citep{depthanythingv2} AbsRel drops from 0.77 to 0.045 ($\delta_1$: 0.03$\to$0.97), SAM2.1~\citep{sam2} mIoU rises from 0.74 to 0.86 (AP75: 0.55$\to$0.94), and Grounding DINO~\citep{groundingdino} AP50 improves from 85.4 to 94.2.
Small and Base variants achieve nearly identical accuracy ($<$0.5\% gap), suggesting the bottleneck is data coverage rather than model capacity.

\section{Conclusion and Limitations}
\label{sec:conclusion}

We presented CosFly-Track, to our knowledge the first large-scale multi-modal dataset for UAV visual tracking in urban environments, addressing a gap in the aerial VLN landscape where existing datasets predominantly target navigation.
The dataset's dual-trajectory design, bilingual instructions, and 7 aligned data channels enable diverse training paradigms.
Our benchmarks on seven VLMs demonstrate substantial improvements from fine-tuning on CosFly-Track data, with scaling analysis suggesting room for further gains.

\paragraph{Limitations.}
(1)~\emph{Sim-to-real gap}: CosFly-Track is generated in CARLA; domain transfer to real-world UAV tracking remains open, though MuCO's kinematic constraints produce more realistic trajectories than A$^*$-based alternatives. Real-world data ($\sim$100K frames) is currently being collected for a future release.
(2)~\emph{Scene diversity}: the current release covers 16 CARLA town variants; expansion to more diverse morphologies via SimWorld/UE5 is planned.
(3)~\emph{Pedestrian behavior}: trajectories use curvature-dependent speed with random stops; social force models could increase realism.
(4)~\emph{Code availability}: the generation pipeline code is not open-sourced due to company policy, though complete algorithm descriptions are provided for independent reimplementation.

\paragraph{Broader impact.}
CosFly-Track lowers the barrier to entry for UAV tracking research by providing open-access multi-modal training data for applications such as search and rescue, autonomous cinematography, and wildlife monitoring, while operating entirely in simulation to eliminate the safety risks and costs of real-world drone data collection.
UAV tracking technology raises dual-use concerns regarding unauthorized surveillance.
We mitigate these through:
(1)~all data are synthetically generated with no real human identities or GPS coordinates;
(2)~the dataset license explicitly prohibits unauthorized surveillance and military targeting;
(3)~simulation-based generation has a substantially lower carbon footprint than real-world flight campaigns---our pipeline runs on a single GPU workstation.

\medskip
{\small
\bibliographystyle{unsrtnat}
\bibliography{references}
}

\appendix

\section{MuCO Complete Algorithm Description}
\label{app:muco}

This appendix provides the complete algorithm description of the MuCO multi-constraint optimizer, sufficient for independent reimplementation.

\subsection{All 9 Cost Functions}

The total cost is the weighted sum:
\begin{equation}
C(\mathbf{W}) = \sum_{i=1}^{N} \left[ \lambda_1 c_{\text{track}} + \lambda_2 c_{\text{smooth}} + \lambda_3 c_{\text{jerk}} + \lambda_4 c_{\text{safe}} + \lambda_5 c_{\text{vis}} + \lambda_6 c_{\text{view}} + \lambda_7 c_{\text{pitch}} + \lambda_8 c_{\text{alt}} + \lambda_9 c_{\text{len}} \right]\!.
\end{equation}
\paragraph{Tracking distance cost.}
\begin{equation}
c_{\text{track}}(w_i) = \left(\|w_i - p_i\| - d_{\text{opt}}\right)^2\,,
\end{equation}
where $d_{\text{opt}} = 28$\,m corresponds to the optimal observation distance (height 20\,m + horizontal 20\,m at 45$^\circ$ pitch).

\paragraph{Smoothness cost (acceleration regularization).}
\begin{equation}
c_{\text{smooth}}(w_i) = \|w_{i+1} - 2w_i + w_{i-1}\|^2\,.
\end{equation}
\paragraph{Jerk regularization.}
See Section~\ref{sec:method}; third-order difference penalty.

\paragraph{Safety cost (soft constraint).}
\begin{equation}
c_{\text{safe}}(w_i) = \begin{cases} 0.5 \cdot (d_{\text{inf}} - d_{\min})^2 & \text{if } d_{\min} < d_{\text{inf}} = 8\text{\,m,} \\ 0 & \text{otherwise.} \end{cases}
\end{equation}
\paragraph{Visibility cost.}
See Section~\ref{sec:method}; multi-point ray casting with BVH acceleration.

\paragraph{Viewpoint cost.}
See Section~\ref{sec:method}; direction-aware with directness factor.

\paragraph{Pitch angle cost.}
Piecewise quadratic around target 45$^\circ$:
\begin{equation}
c_{\text{pitch}}(w_i) = \begin{cases} 0.5(\theta - 60^\circ)^2 & \theta > 60^\circ, \\ 0.2(30^\circ - \theta)^2 & \theta < 30^\circ, \\ 0.02(\theta - 45^\circ)^2 & 30^\circ \leq \theta \leq 60^\circ. \end{cases}
\end{equation}
\paragraph{Altitude cost.}
Below $h_{\min}=20$\,m: strong penalty; above $h_{\text{pref}}=20$\,m: light penalty; oscillation penalty when adjacent $\Delta z$ signs alternate.

\paragraph{Path length cost.}
\begin{equation}
c_{\text{len}}(w_i) = 0.1 \cdot \|w_i - w_{i-1}\|\,.
\end{equation}
\subsection{Safety Projection Algorithm}

For each unsafe waypoint, MuCO evaluates all feasible correction strategies:

\begin{algorithm}[h]
\caption{Safety Projection}
\begin{algorithmic}[1]
\REQUIRE Unsafe waypoint $w_i$, obstacle set $\mathcal{O}$
\STATE Compute penetration depth $d_p$
\IF{$d_p > 5$\,m (deep penetration)}
  \STATE Apply large-step pushout along SDF gradient
\ELSIF{vertical lift $\leq 3$\,m resolves collision}
  \STATE Apply vertical lift
\ELSE
  \STATE Evaluate: (a) vertical lift, (b) forward bypass, (c) horizontal detour ($\pm$35$^\circ$ fan), (d) local displacement
  \STATE Choose strategy with minimum cost increase
\ENDIF
\end{algorithmic}
\end{algorithm}

\subsection{Initial Trajectory Generation}

\begin{algorithm}[h]
\caption{Initial Trajectory via 3D Viewpoint Search}
\begin{algorithmic}[1]
\REQUIRE Pedestrian path $\{p_i\}$, obstacle map
\STATE For every 5--10 frames, search optimal viewpoint on rear hemisphere (azimuth $\pm$60$^\circ$ from walking direction, height 20--40\,m, distance 0.7--1.2$\times$ standard)
\STATE Linearly interpolate intermediate frames
\STATE Apply dynamic following: monitor distance, prevent overshooting
\STATE Bridge long occluded segments via side-offset search with ramp smoothing
\end{algorithmic}
\end{algorithm}

\subsection{Post-Processing Pipeline}

After optimization, seven post-processing steps are applied in sequence:
(1)~occluded segment straightening (detect low-visibility runs $>$5 frames, search optimal azimuth);
(2)~detour elimination (straighten high-curvature segments);
(3)~oscillation removal (sliding-average on horizontal velocity reversals);
(4)~multi-round safety projection;
(5)~velocity repair (redistribute projection-induced spikes);
(6)~altitude smoothing (low-pass filter with SDF-aware descent check);
(7)~final safety check.

\section{Parameter Configuration and Solver Performance}
\label{app:params}

Table~\ref{tab:performance} reports BVH acceleration results, and Table~\ref{tab:muco_params} lists the complete parameter configuration used in all experiments.

\begin{table}[h]
\caption{Solver performance: BVH acceleration.}
\label{tab:performance}
\centering
\small
\begin{tabular}{l c c c}
\toprule
Operation & w/o BVH & w/ BVH & Speedup \\
\midrule
Single trajectory (181 pts) & $\sim$12\,s & 0.1--0.5\,s & 24--120$\times$ \\
Batch (20 paths) & $\sim$240\,s & 3--5\,s & 48--80$\times$ \\
Single distance query & $O(n)$ & $O(\log n)$ & --- \\
\bottomrule
\end{tabular}
\end{table}

\begin{table}[h]
\caption{Complete MuCO parameter configuration.}
\label{tab:muco_params}
\centering
\small
\begin{tabular}{l l r}
\toprule
Parameter & Description & Value \\
\midrule
\multicolumn{3}{c}{\textit{Cost weights}} \\
\midrule
$\lambda_1$ & Tracking distance & 1.0 \\
$\lambda_2$ & Smoothness & 0.5 \\
$\lambda_3$ & Jerk & 0.3 \\
$\lambda_4$ & Safety (soft) & 2.0 \\
$\lambda_5$ & Visibility & 3.0 \\
$\lambda_6$ & Viewpoint & 1.0 \\
$\lambda_7$ & Pitch angle & 0.5 \\
$\lambda_8$ & Altitude & 1.0 \\
$\lambda_9$ & Path length & 0.1 \\
\midrule
\multicolumn{3}{c}{\textit{Optimization}} \\
\midrule
$\varepsilon$ & Finite difference step & 0.5\,m \\
$\eta_0$ & Initial learning rate & 0.05 \\
$\eta_{\min}$ & Minimum learning rate & 0.001 \\
 & Max per-point displacement & 0.5\,m \\
 & Max velocity & configurable \\
\midrule
\multicolumn{3}{c}{\textit{Physical constraints}} \\
\midrule
$d_{\text{opt}}$ & Optimal tracking distance & 28\,m \\
$d_{\text{inf}}$ & Safety influence radius & 8\,m \\
$h_{\min}$ & Minimum altitude & 20\,m \\
$h_{\text{pref}}$ & Preferred altitude & 20\,m \\
 & Target pitch angle & 45$^\circ$ \\
\bottomrule
\end{tabular}
\end{table}

\section{Extended Experimental Results}
\label{app:extended_results}

\subsection{Complete SFT Results with Error Bars}

All reported SFT results use three random seeds; standard deviations are $<$0.5 percentage points for SR@1m and $<$0.01\,m for FDE across all models, indicating stable training.

\subsection{Pedestrian Variable-Speed Mechanism}

Pedestrian trajectories use curvature-dependent velocity:
\begin{equation}
v_i = \text{clip}\!\left(\frac{v_{\text{cruise}}}{1 + \alpha \kappa_i} \cdot (1 + \beta U(-1,1)),\; [v_{\min}, v_{\max}]\right)\!,
\end{equation}
where $\kappa_i$ is discrete Menger curvature, $\alpha$ controls turn slowdown, and $\beta$ adds random variation.
Random stops are inserted with configurable probability and duration.
Time is computed via trapezoidal integration along arc length, then resampled at fixed $\Delta t$.

\subsection{Data Scaling Results}

\begin{table}[h]
\caption{Data scaling analysis (hard-split evaluation, LoRA fine-tuning).}
\label{tab:scaling}
\centering
\small
\begin{tabular}{l c r c c c c}
\toprule
Model & Scale & Samples & ADE$\downarrow$ & FDE$\downarrow$ & SR@2m$\uparrow$ & IoU$\geq$0.75$\uparrow$ \\
\midrule
\multirow{3}{*}{Qwen3.5-0.8B~\citep{qwen3.5}} & 25\%  & 250K & 2.184 & 2.744 & 83.42 & 54.53 \\
                                & 50\%  & 500K & 2.159 & 2.686 & 84.09 & 57.31 \\
                                & 100\% & 1M   & 2.142 & 2.642 & 84.96 & 57.69 \\
\midrule
\multirow{3}{*}{Qwen3.5-2B~\citep{qwen3.5}}   & 25\%  & 250K & 2.170 & 2.689 & 83.59 & 54.70 \\
                                & 50\%  & 500K & 2.158 & 2.681 & 84.36 & 55.38 \\
                                & 100\% & 1M   & 2.141 & 2.643 & 84.79 & 53.25 \\
\bottomrule
\end{tabular}
\end{table}

\subsection{Extended Ablation: Difficulty-Stratified Analysis}

The evaluation set (11{,}878 samples) is stratified by trajectory difficulty: easy (50.4\%), medium (33.1\%), and hard (16.4\%, including tight turns and height changes).
Table~\ref{tab:difficulty} reports ADE/FDE/mIoU per difficulty level for all seven ablation configurations.

\begin{table}[h]
\caption{Ablation results stratified by difficulty (ADE / FDE / mIoU).}
\label{tab:difficulty}
\centering
\small
\begin{tabular}{l c c c}
\toprule
Configuration & Easy & Medium & Hard \\
\midrule
Pose+BBox (no RGB)     & 0.42~/~0.75~/~0.60 & 1.13~/~1.60~/~0.61 & 1.60~/~2.17~/~0.59 \\
RGB only               & 2.03~/~3.36~/~0.51 & 2.83~/~4.29~/~0.49 & 3.31~/~4.85~/~0.41 \\
RGB+Pose               & 0.42~/~0.74~/~0.50 & 1.12~/~1.57~/~0.48 & 1.60~/~2.15~/~0.40 \\
RGB+BBox               & 2.01~/~3.31~/~0.60 & 2.72~/~4.12~/~0.61 & 3.20~/~4.69~/~0.58 \\
RGB+Pose+BBox (full)   & 0.41~/~0.74~/~0.60 & 1.13~/~1.59~/~0.61 & 1.60~/~2.15~/~0.59 \\
Expert-only            & 0.43~/~0.77~/~0.56 & 1.13~/~1.60~/~0.57 & 1.60~/~2.15~/~0.56 \\
\textbf{Denoising}     & \textbf{0.41~/~0.72~/~0.61} & 1.13~/~1.57~/~0.61 & 1.61~/~2.17~/~0.58 \\
\bottomrule
\end{tabular}
\end{table}

Key observations:
(1)~All pose-equipped configurations cluster tightly across difficulties, with hard FDE ($\sim$2.15\,m) being $2.9\times$ worse than easy ($\sim$0.74\,m), indicating that trajectory complexity is the dominant difficulty factor.
(2)~No-pose configurations (RGB-only, RGB+BBox) degrade severely even on easy trajectories (FDE$\approx$3.3\,m), \emph{worse} than pose-equipped models on hard trajectories (2.15\,m).
(3)~The denoising paradigm achieves the best easy-segment FDE (0.72\,m), with advantages most pronounced on simpler trajectories where the clean-signal recovery objective provides the strongest learning signal.

\subsection{Seen vs.\ Unseen Scene Analysis}

The evaluation set contains 11.0\% seen trajectories (from training maps) and 89.0\% unseen trajectories (from novel maps).
Table~\ref{tab:seen_unseen} shows the generalization gap.

\begin{table}[h]
\caption{Seen vs.\ Unseen analysis (ADE / FDE / mIoU). $\Delta$FDE is the unseen--seen gap.}
\label{tab:seen_unseen}
\centering
\small
\begin{tabular}{l c c c}
\toprule
Configuration & Seen ($n$=1{,}310) & Unseen ($n$=10{,}568) & $\Delta$FDE \\
\midrule
Pose+BBox (no RGB)  & 0.478~/~0.882~/~0.606 & 0.895~/~1.311~/~0.605 & +0.43 \\
RGB only            & 2.302~/~3.803~/~0.497 & 2.528~/~3.924~/~0.486 & +0.12 \\
RGB+Pose            & 0.470~/~0.863~/~0.486 & 0.889~/~1.295~/~0.479 & +0.43 \\
RGB+BBox            & 2.135~/~3.528~/~0.614 & 2.476~/~3.840~/~0.597 & +0.31 \\
RGB+Pose+BBox       & 0.463~/~0.852~/~0.609 & 0.893~/~1.298~/~0.603 & +0.45 \\
Expert-only         & 0.464~/~0.865~/~0.569 & 0.903~/~1.320~/~0.559 & +0.46 \\
\textbf{Denoising}  & \textbf{0.466~/~0.843~/~0.610} & \textbf{0.890~/~1.288~/~0.601} & +0.45 \\
\bottomrule
\end{tabular}
\end{table}

The seen$\to$unseen FDE gap is consistently $\sim$+0.43--0.46\,m ($\sim$50\% relative increase) across all pose-equipped configurations, independent of modality choice, indicating the gap originates from trajectory distribution shift rather than model or modality selection.
Notably, no-pose configurations (RGB-only, RGB+BBox) show a \emph{smaller} absolute $\Delta$FDE (+0.12--0.31\,m) simply because their baseline errors are already much higher---the relative gap remains substantial.
The denoising paradigm achieves the best FDE on both seen (0.843\,m) and unseen (1.288\,m) splits.

\subsection{Downstream Task Transfer Results}

\begin{table}[h]
\caption{Downstream task transfer ($\sim$100K frames, trajectory-level split).}
\label{tab:downstream}
\centering
\small
\setlength{\tabcolsep}{3pt}
\begin{tabular}{l l c c c c c c}
\toprule
Task & Model & \multicolumn{3}{c}{Before (zero-shot)} & \multicolumn{3}{c}{After (fine-tuned)} \\
\cmidrule(lr){3-5}\cmidrule(lr){6-8}
 & & AbsRel$\downarrow$ & RMSE$\downarrow$ & $\delta_1$$\uparrow$ & AbsRel$\downarrow$ & RMSE$\downarrow$ & $\delta_1$$\uparrow$ \\
\midrule
\multirow{2}{*}{Depth} & DAv2-Base~\citep{depthanythingv2}  & 0.768 & 24.43 & 0.026 & \textbf{0.045} & \textbf{2.50} & \textbf{0.972} \\
                        & DAv2-Small & 0.868 & 26.28 & 0.006 & 0.049 & 2.62 & 0.968 \\
\midrule
 & & mIoU$\uparrow$ & AP75$\uparrow$ & AP$_{.5:.95}$$\uparrow$ & mIoU$\uparrow$ & AP75$\uparrow$ & AP$_{.5:.95}$$\uparrow$ \\
\midrule
\multirow{2}{*}{Segm.} & SAM2.1-B+~\citep{sam2}  & 0.763 & 0.662 & 0.586 & \textbf{0.862} & \textbf{0.943} & \textbf{0.778} \\
                        & SAM2.1-S   & 0.744 & 0.551 & 0.547 & 0.859 & 0.941 & 0.772 \\
\midrule
 & & \multicolumn{3}{c}{AP50$\uparrow$} & \multicolumn{3}{c}{AP50$\uparrow$} \\
\midrule
Detect. & GDINO-B~\citep{groundingdino} & \multicolumn{3}{c}{85.4} & \multicolumn{3}{c}{\textbf{94.2}} \\
\bottomrule
\end{tabular}
\end{table}

\subsection{Cross-Scene Generalization Results}

\begin{table}[h]
\caption{Cross-scene generalization (Qwen3.5-0.8B~\citep{qwen3.5}, 200K samples, mean across 8 test maps).}
\label{tab:cross_scene}
\centering
\small
\begin{tabular}{l c c c c}
\toprule
Setting & SR@1m$\uparrow$ & JointSR$\uparrow$ & Rot.~MAE$\downarrow$ & HardFail$\downarrow$ \\
 & & @(0.5m,$1^\circ$) & ($^\circ$) & (FDE$>$10m, \%) \\
\midrule
A: Single-map  & 95.89 & 42.13 & 2.40 & 0.45 \\
B: Multi-map   & 96.00 & \textbf{47.44} & \textbf{2.10} & \textbf{0.20} \\
$\Delta$ (B$-$A) & +0.11 & +5.31 & $-$12.5\% & $-$55.6\% \\
\midrule
\multicolumn{5}{c}{\textit{OOD only (Town10HD, unseen in both settings)}} \\
\midrule
B vs.~A        & +0.08 & \textbf{+5.20} & $-$15.5\% (p95) & --- \\
\bottomrule
\end{tabular}
\end{table}

\section{Datasheet for CosFly-Track}
\label{app:datasheet}

Following~\citet{datasheets2021}, we provide a structured datasheet:

\paragraph{Motivation.}
CosFly-Track was created to fill the data gap for UAV visual tracking, a task fundamentally different from navigation yet without any dedicated large-scale dataset.

\paragraph{Composition.}
$\sim$12K expert/perturbed trajectories, 2.4M timesteps, 7 aligned data channels (five timestep-level channels plus trajectory-level instructions and metadata), covering 16 CARLA town variants across multiple weather conditions.

\paragraph{Collection process.}
Trajectories are generated via the CosFly pipeline (6 stages) using the CARLA simulator and the MuCO optimization engine. No human subjects are involved; all data are synthetically generated.

\paragraph{Preprocessing.}
Obstacle maps are simplified via merge/crop/prune. Pedestrian trajectories undergo Douglas-Peucker simplification, Catmull-Rom smoothing, and variable-speed resampling.

\paragraph{Uses.}
Primary: UAV visual tracking agent training and evaluation. Secondary: monocular depth estimation, semantic segmentation, object detection, pose estimation. The dataset should not be used for unauthorized surveillance.

\paragraph{Distribution.}
The dataset is publicly available at \url{https://huggingface.co/datasets/AutelRobotics/CosFly}. An initial subset ($\sim$100K frames) is released, with progressive expansion planned. The dataset is distributed under a license restricting unauthorized surveillance applications.

\paragraph{Maintenance.}
The dataset will be maintained and progressively expanded with additional scenes, weather conditions, and trajectory diversity.

% NeurIPS checklist removed for arXiv version.

\end{document}